\documentclass[11pt]{article}

\usepackage[preprint]{acl}

\usepackage{times}
\usepackage{latexsym}

\usepackage[T1]{fontenc}

\usepackage[utf8]{inputenc}

\usepackage{microtype}

\usepackage{inconsolata}

\usepackage{graphicx}

\usepackage{microtype}
\usepackage{booktabs}
\usepackage{authblk}

\usepackage[utf8]{inputenc}   
\usepackage[T1]{fontenc}            
\usepackage{url}              
\usepackage{nicefrac}         
\usepackage{lipsum}           
\usepackage{natbib}           
\usepackage{doi}              
\usepackage{array}            
\usepackage{multirow}         
\usepackage{siunitx}          
\usepackage{caption}          
\usepackage{amsmath}          
\usepackage{wrapfig}          
\usepackage{pifont}           
\usepackage{tikz}             
\usepackage{float}                           
\usepackage{overpic}  
\usepackage{graphicx}   
\usepackage{subfig} 
\usepackage{xcolor}   
\usepackage{enumitem} 
\usepackage{hyperref}
\usepackage{amsmath}
\usepackage{amssymb}

\usepackage{booktabs}
\usepackage{multirow}
\usepackage{graphicx}
\usepackage[table]{xcolor}
\usepackage{arydshln}
\usepackage{pifont}

\usepackage[most]{tcolorbox}
\usepackage{xcolor}

\usepackage{hyperref}
\usepackage{fontawesome5}

\tcbset{
promptbox/.style={
    enhanced,
    breakable,
    colframe=mitred,
    colback=mitred!3,
    opacityback=0.95,
    boxrule=0.7pt,
    arc=4pt,
    left=6pt,
    right=6pt,
    top=5pt,
    bottom=5pt,
    fonttitle=\bfseries,
    coltitle=white,
    colbacktitle=mitred,
    boxed title style={
        arc=3pt,
        boxrule=0pt,
        left=5pt,
        right=5pt,
        top=2pt,
        bottom=2pt
    },
    attach boxed title to top left={xshift=6pt,yshift=-2pt},
}
}

\lstdefinestyle{promptjson}{
    basicstyle=\ttfamily\scriptsize,
    breaklines=true,
    columns=fullflexible,
    keepspaces=true,
    showstringspaces=false,
    frame=none,
    xleftmargin=0pt,
    xrightmargin=0pt
}

\definecolor{mitred}{RGB}{160,30,45} 

\newcommand{\cmark}{\ding{51}}
\newcommand{\xmark}{\ding{55}}

\definecolor{graybg}{RGB}{242,242,242}
\definecolor{crimson}{RGB}{220,20,60}
\definecolor{forestgreen}{RGB}{34,139,34}

\title{Text-Only Data Synthesis for Vision Language Model Training}

\author{
  \textbf{Xiaomin Yu\textsuperscript{1}},
  \textbf{Wenjie Zhang\textsuperscript{1}},
  \textbf{Ziyue Qiao\textsuperscript{2}},
  \textbf{Chengwei Qin\textsuperscript{1,*}},
  \textbf{Hui Xiong\textsuperscript{1}}
\\
\\
  \textsuperscript{1} The Hong Kong University of Science and Technology (Guangzhou)
\\
  \textsuperscript{2} Great Bay University
  \\
    \\
  \small{
  \faGithub\ \textbf{GitHub:} \url{https://github.com/Yu-xm/Modality_Gap_Theory.git}
}
\\
  \small{
    \textbf{Email:} \href{mailto:yuxm02@gmail.com}{yuxm02@gmail.com}
  }
  \\
  \small{
    \textsuperscript{*} Corresponding Author.
  }
}

\begin{document}
\maketitle

\def\method{Unicorn}

\begin{abstract}
Training vision-language models (VLMs) typically requires large-scale, high-quality image-text pairs, but collecting or synthesizing such data is costly. In contrast, text data is abundant and inexpensive, prompting the question: can high-quality multimodal training data be synthesized purely from text? To address this problem, we propose a cross-integrated three-stage multimodal data synthesis framework, termed Unicorn, which generates two datasets: Unicorn-1.2M and Unicorn-471K-Instruction. In Stage 1: Diverse Caption Data Synthesis, we construct 1.2M semantically diverse high-quality captions by expanding sparse caption seeds using large language models (LLMs). In Stage 2: Instruction-Tuning Data Generation, we further process 471K captions into multi-turn instruction-tuning tasks to support complex reasoning. Finally, in Stage 3: Modality Representation Transfer, these textual captions representations are transformed into visual representations, resulting in diverse synthetic image representations. This three-stage process enables us to construct Unicorn-1.2M for pretraining and Unicorn-471K-Instruction for instruction-tuning, without relying on real images. By eliminating the dependency on real images while maintaining data quality and diversity, our framework offers a cost-effective and scalable solution for VLM training.
\end{abstract}

\begin{figure}[ht!]
  \centering
  \includegraphics[width=0.5\textwidth]{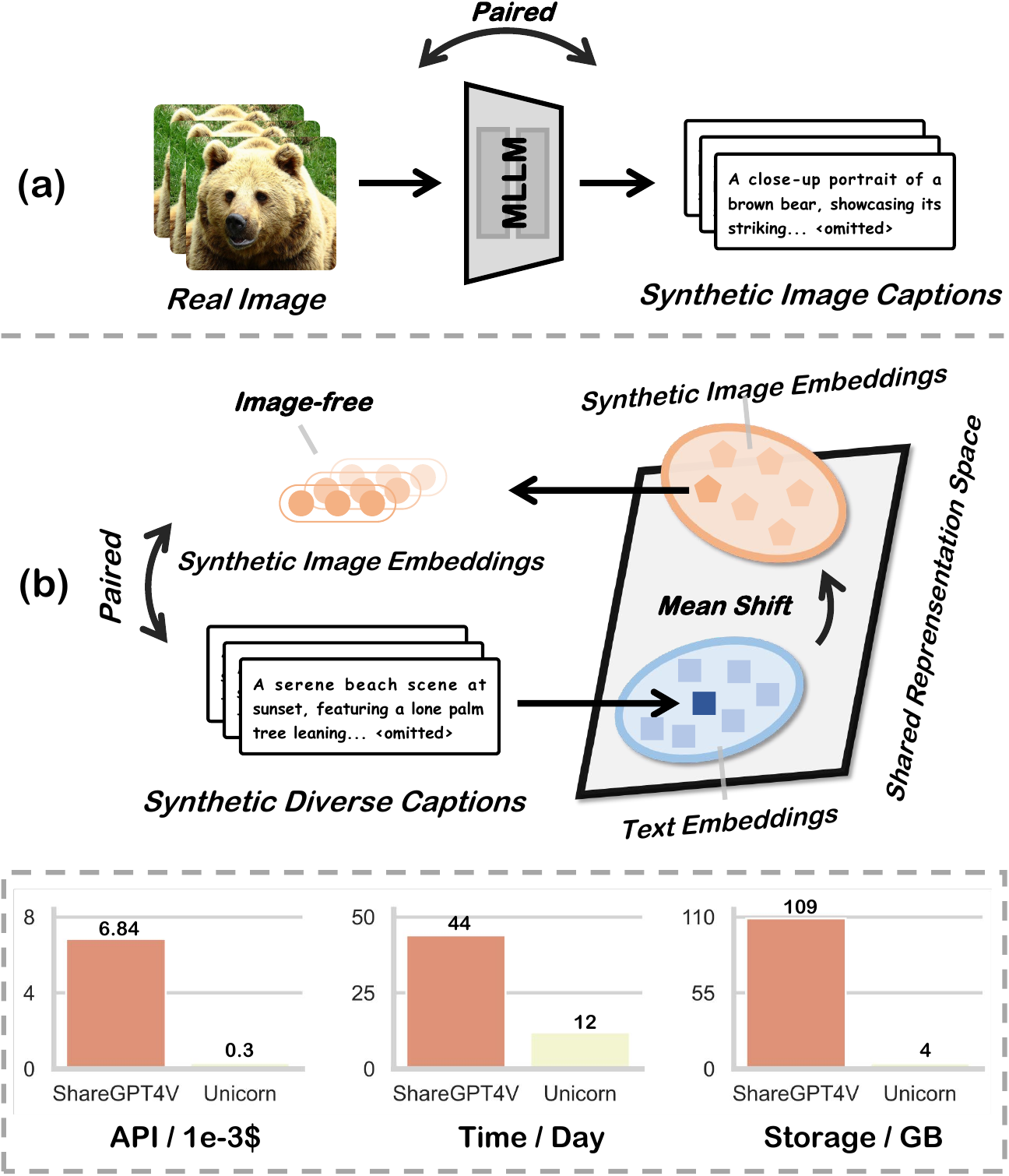}
  \caption{Unlike traditional image-text data synthesis frameworks, \method{} removes the dependency on real image data, offering a more efficient solution.}
  \label{fig:intro}
  \vspace{-8pt}
\end{figure}

\section{Introduction}

Vision-Language Models (VLMs) \cite{liu2024improved,he2024efficient} have become a central paradigm for bridging visual and textual modalities, enabling models to understand, generate, and reason over multimodal inputs. Recent VLMs have demonstrated strong capabilities across visual question answering, image captioning, multimodal reasoning, and human-computer interaction. However, the success of high-performance VLMs largely depends on large-scale, high-quality, and diverse image-text datasets. Constructing such datasets remains costly and difficult, as it requires not only massive amounts of multimodal data but also careful curation to ensure data quality, semantic diversity, and cross-modal alignment.

Existing approaches for acquiring image-text training data mainly rely on two strategies. Manual annotation \cite{lin2014microsoft,plummer2015flickr30k} provides high-quality captions but is limited by annotation cost, dataset scale, and semantic diversity. Large-scale web crawling \cite{sharma2018conceptual} improves scalability, but introduces noisy captions, inconsistent data quality, compliance risks, and potential copyright concerns. Recent works \cite{chen2024sharegpt4v,li2024densefusion} augment real images with model-generated captions, such as fine-grained captions produced by powerful multimodal models. Although these methods improve data quality, they still require real images and expensive multimodal generation pipelines.

In contrast, text data is abundant, inexpensive, and highly diverse. This raises a fundamental question: \textbf{Can text-only data provide effective multimodal supervision for VLM training without relying on real or synthetic images?} More specifically, we ask whether visual supervision can be constructed by text directly in a shared representation space, as shown in Fig.~\ref{fig:intro}. A key obstacle to this idea is the modality gap \cite{zhang2024connect}: even in a shared representation space, text embeddings and image embeddings with similar semantics are not perfectly aligned, but are separated by systematic modality-dependent biases. Therefore, if textual representations can be transformed into visually compatible representations by mitigating this modality gap, then text-only data may serve as a scalable source of multimodal supervision \cite{tewel2022zerocap,gu2023can,liu2024arcsin}.

To investigate this possibility, we first conduct a large-scale analysis of paired image-text representations. Our analysis reveals three empirical findings: \ding{182} the image-text modality gap contains a stable global component; \ding{183} this component is largely consistent across semantic regions; and \ding{184} it is transferable across unseen modality regions. These findings suggest that text-derived representations can be transformed into visually compatible representations through a shared transfer rule.

Motivated by these findings, we propose \textbf{Unicorn}, a scalable text-only multimodal data synthesis framework. Unicorn synthesizes data in the representation space. Specifically, Unicorn first expands sparse textual seeds into dense visual captions, then derives instruction-following data from these captions, and finally transfers their text embeddings into synthetic image representations in the shared embedding space. This pipeline constructs two datasets without using real images: Unicorn-1.2M for pretraining and Unicorn-471K-Instruction for instruction tuning.

Using these datasets, we train \textbf{Unicorn-8B}, a VLM trained entirely from text-only synthetic multimodal supervision. Experimental results show that Unicorn-8B achieves competitive performance across multiple benchmarks compared with methods trained on conventional image-text pairs, demonstrating the feasibility of large-scale VLM training without real images.


Overall, our contributions are threefold. \ding{182} We reveal a stable and transferable modality gap structure. \ding{183} We propose \method{}, a text-only data synthesis framework for VLM Training. \ding{184} We show that text-only synthetic data can effectively train a competitive VLM without real images.
\section{Modality Gap Analysis} \label{sec2}

\begin{figure*}[t]
    \centering
    \includegraphics[width=\textwidth]{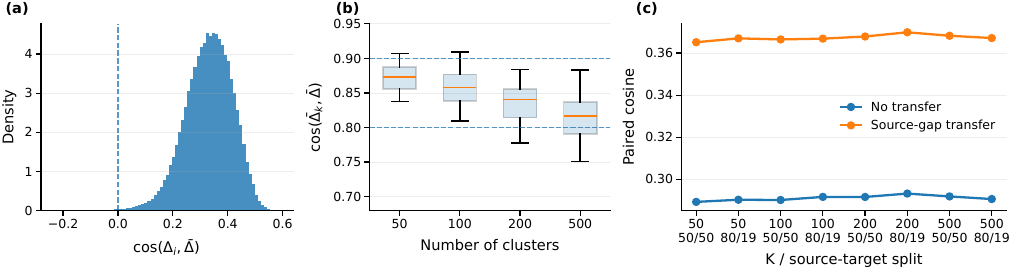}
    \caption{
    Empirical principles of the modality gap in the shared representation space.
    (a) Pairwise modality gaps are directionally aligned with the global gap, indicating a stable global modality bias.
    (b) Local gaps estimated from unsupervised semantic clusters remain consistently aligned with the global gap across different clustering granularities.
    (c) A gap estimated from source clusters consistently improves paired cosine similarity on disjoint target clusters, demonstrating cross-region transferability.
    }
    \label{fig2}
\end{figure*}

We first examine whether text representations can be made visually compatible in a shared representation space. Since text and image embeddings may remain systematically shifted even with similar semantics, directly treating text embeddings as image embeddings is suboptimal. We therefore analyze paired image-text representations to characterize the modality gap. Given 1M image-text pairs $\{(x_i,y_i)\}_{i=1}^{1M}$, we use the image encoder $f_I(\cdot)$ and text encoder $f_T(\cdot)$ of LLM2CLIP to map images and texts into the same shared representation space, i.e., $v_i=f_I(x_i)$ and $u_i=f_T(y_i)$, where $v_i,u_i\in\mathbb{R}^{d}$. All representations are $\ell_2$-normalized before computation. For each image-text pair, we define the pairwise modality gap as $\Delta_i=v_i-u_i$, and define the global modality gap as $\bar{\Delta}=N^{-1}\sum_i\Delta_i$. The pairwise gap can then be decomposed as:
\begin{equation}
v_i=u_i+\bar{\Delta}+\epsilon_i,
\end{equation}
where $\epsilon_i$ denotes the residual alignment noise after removing the global modality bias. 


\paragraph{Finding 1: Stable Global Bias.} We first analyze the directional consistency between the pairwise gap $\Delta_i$ and the global gap $\bar{\Delta}$. As shown in Fig.~\ref{fig2}(a), the distribution of $\cos(\Delta_i,\bar{\Delta})$ is mostly concentrated in the positive region, with a peak around $0.35$, and the average directional consistency is approximately $0.33$. This indicates that most pairwise gaps are positively aligned with the global gap, suggesting the existence of a stable global modality-bias direction. We further compute the gap energy concentration ratio $\rho=\|\bar{\Delta}\|_2^2 / (N^{-1}\sum_i\|\Delta_i\|_2^2)$. In our experiments, $\rho\approx0.11$, indicating that the global gap explains approximately $11\%$ of the pairwise gap energy. Thus, the global component is a stable and exploitable structural component. Additional analyses in the Appendix~\ref{App:B} show that the residual term $\epsilon_i=\Delta_i-\bar{\Delta}$ remains substantial but is approximately zero-centered. These results suggest that image-text discrepancy is better modeled as a stable global bias plus residual noise.


\begin{tcolorbox}[colframe=mitred, opacityback=0.9, arc=4pt, left=4pt, right=4pt, top=2pt, bottom=0pt,
title=Implication 1.]
This finding suggests that text representations should not be directly used as image representations. Instead, a representation-space transfer step is required to compensate for the global modality bias.
\end{tcolorbox}


\paragraph{Finding 2: Cross-Semantic Consistency.}
We next examine whether the global gap is strongly tied to specific semantics. Since the 1M paired data do not contain domain annotations, we approximate semantic regions in an unsupervised manner. Specifically, we apply K-means clustering to text representations $u_i$ and treat each cluster as a local semantic region, with $K\in\{50,100,200,500\}$. For the $k$-th semantic cluster $\mathcal{C}_k$, we define its local gap as $\bar{\Delta}_k=|\mathcal{C}_k|^{-1}\sum_{i\in\mathcal{C}_k}(v_i-u_i)$, and compute its directional consistency with the global gap, namely $\cos(\bar{\Delta}_k,\bar{\Delta})$. As shown in Fig.~\ref{fig2}(b), local gaps remain strongly aligned with the global gap across different clustering granularities. Most cluster-level similarities are concentrated in the range of approximately $0.8$ to $0.9$, and the distribution only decreases mildly as $K$ increases. This trend is expected because larger $K$ produces finer semantic clusters with fewer samples per cluster, making local gap estimates more sensitive to sample-level residuals. Nevertheless, even under the finest clustering setting, local gaps remain substantially and positively aligned with the global gap. These results indicate that the dominant modality-gap direction is not specific to individual semantic categories, but mainly reflects a shared modality-level bias across semantic regions.

\begin{tcolorbox}[colframe=mitred, opacityback=0.9, arc=4pt, left=4pt, right=4pt, top=2pt, bottom=0pt,
title=Implication 2.]
This finding suggests that a shared transfer rule can be applied to diverse textual data, rather than requiring a separate modality correction for every semantic category.
\end{tcolorbox}

\begin{figure*}[h!]
  \centering
  \includegraphics[width=\textwidth]{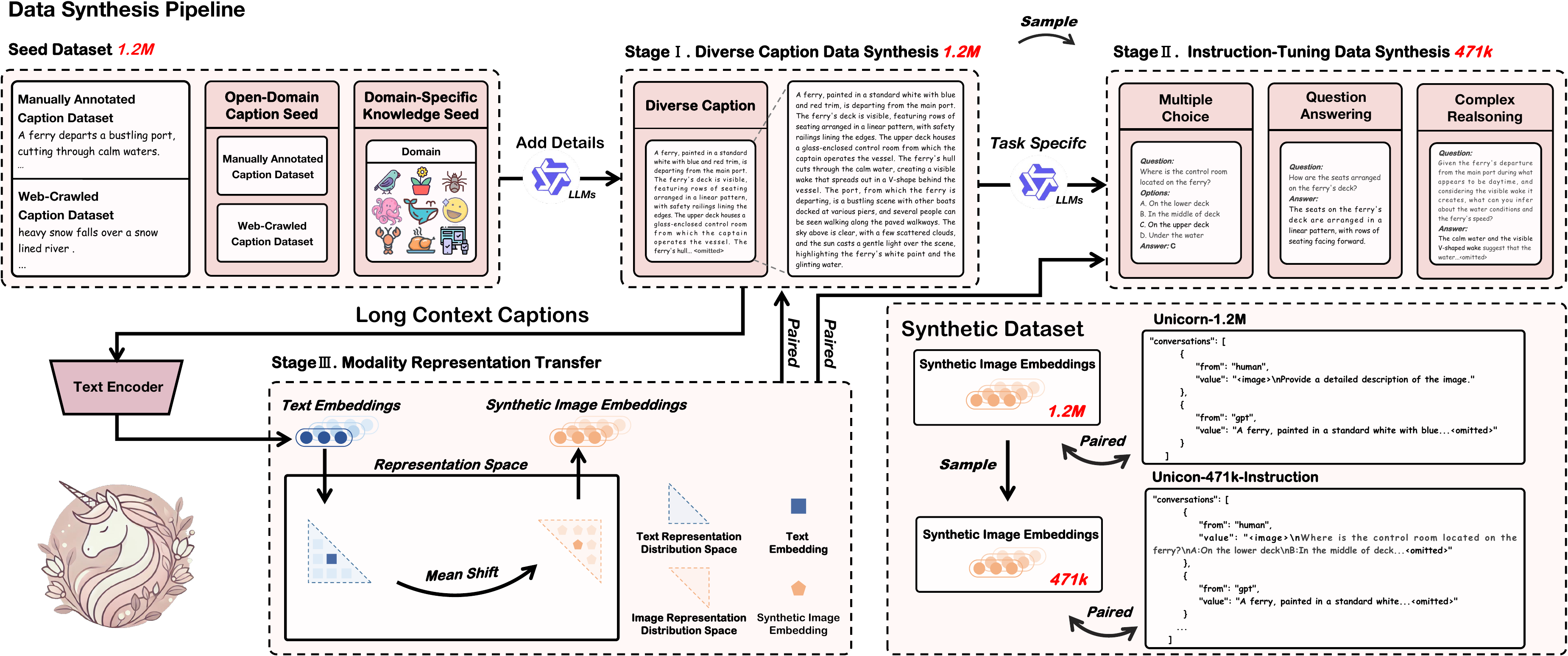}
  \caption{\method{}’s text-only data synthesis pipeline, comprising three cross-integrated stages, (1) Diverse Caption Data Synthesis, (2) Instruction-Tuning Data Synthesis, and (3) Representation Transfer, ultimately yields two synthetic datasets: Unicorn-1.2M and Unicorn-471K-Instruction. Notably, unlike traditional approaches that rely on real images, \method{} generates image representations paired with text, entirely free of real image data.}
  \label{fig:pipeline}
\end{figure*}

\paragraph{Finding 3: Cross-Modality Transferability.} Directional consistency alone does not guarantee transferability. We therefore conduct a cross-cluster gap transfer experiment. Based on the text-clustering results, we split clusters into source clusters and target clusters. We estimate the gap only using image-text pairs from source clusters, i.e., $\hat{c}_{src}=|\mathcal{D}_{src}|^{-1}\sum_{i\in\mathcal{D}_{src}}(v_i-u_i)$, and apply it to disjoint target clusters by constructing $\hat{v}_i=u_i+\hat{c}_{src}$. On target clusters, we compare $\cos(u_i,v_i)$ before transfer with $\cos(u_i+\hat{c}_{src},v_i)$ after transfer. As shown in Fig.~\ref{fig2}(c), under all $K$ values and source/target split settings, the source-estimated gap consistently improves paired cosine similarity on target clusters. Specifically, the paired cosine similarity before transfer is around $0.29$, and increases to approximately $0.365\sim0.370$ after transfer, with an average improvement of about $+0.075$. This improvement remains stable across different clustering granularities and source/target split ratios. This result provides direct evidence that the modality gap contains a transferable component shared across semantic regions.


\begin{tcolorbox}[colframe=mitred, opacityback=0.9, arc=4pt, left=4pt, right=4pt, top=2pt, bottom=0pt,
title=Implication 3.]
This finding shows that the modality gap structure can generalize to unseen semantic regions. Therefore, for synthetic captions without paired images, we can still apply a shared transfer rule to convert text embeddings into visually compatible representations.
\end{tcolorbox}

\section{Data Synthesis Pipeline}

Motivated by the empirical principles established in Sec.~\ref{sec2}, we design Unicorn as a text-only data synthesis pipeline for constructing visually compatible multimodal supervision. The key idea is to first make textual captions sufficiently visualizable, then convert their dense visual semantics into instruction-following supervision, and finally transfer their text embeddings toward the visual representation space through modality-gap-guided representation transfer.

\subsection{Seed Dataset}

According to the findings in Sec.~\ref{sec2}, the dominant modality gap remains largely consistent across semantic regions, suggesting that a unified representation transfer rule can be applied to diverse textual sources. Therefore, before synthesis, we construct a seed dataset that covers both open-domain visual scenarios and domain-specific concepts.

To construct high-quality, diverse captions, we collect a seed dataset with two main objectives: \ding{182} to provide diverse textual visual scenarios for fine-grained alignment during VLM pretraining, and \ding{183} to incorporate domain-specific knowledge, enhancing the model's understanding of specialized content. Importantly, we use only the textual captions from these datasets without involving any images. Specifically, given $M$ caption datasets $\{D_m\}_{m=1}^{M}$, where each dataset $D_m$ contains image captions $\{s_i\}_{i=1}^{N_m}$, we filter and merge the original textual captions from these datasets, resulting in a seed dataset $S=\{s_1,s_2,\ldots,s_N\}$ containing 1.2M high-quality textual samples, referred to as Unicorn-1.2M-Seed. This seed dataset consists of two categories: open-domain caption seeds and domain-specific knowledge seeds.

\noindent \textbf{Open-Domain Caption Seeds.} Open-domain seeds include two types of image caption datasets: manually annotated captions and web-crawled captions. Manually annotated caption datasets provide basic information about entities, events, and contexts. To improve diversity and reduce redundancy, we retain only one caption per image. Web-crawled caption datasets provide large-scale and semantically diverse captions. Since web-crawled captions may contain noise, we filter and select captions that more closely resemble high-quality captions.

\noindent \textbf{Domain-Specific Knowledge Seeds.} To inject domain-specific knowledge, we collect 10K fine-grained image labels from six fine-grained image classification datasets across 18 domains. For each dataset, we design specialized prompt templates. For example, we use templates such as ``An image of a delicious \texttt{<label>}'', where \texttt{<label>} is dynamically replaced with the actual class name. Each label is paired with 10 randomly selected templates. This process generates 107K domain-specific knowledge seeds, enriching the seed dataset with fine-grained domain concepts.

\begin{figure*}[h!]
  \centering
  \includegraphics[width=\textwidth]{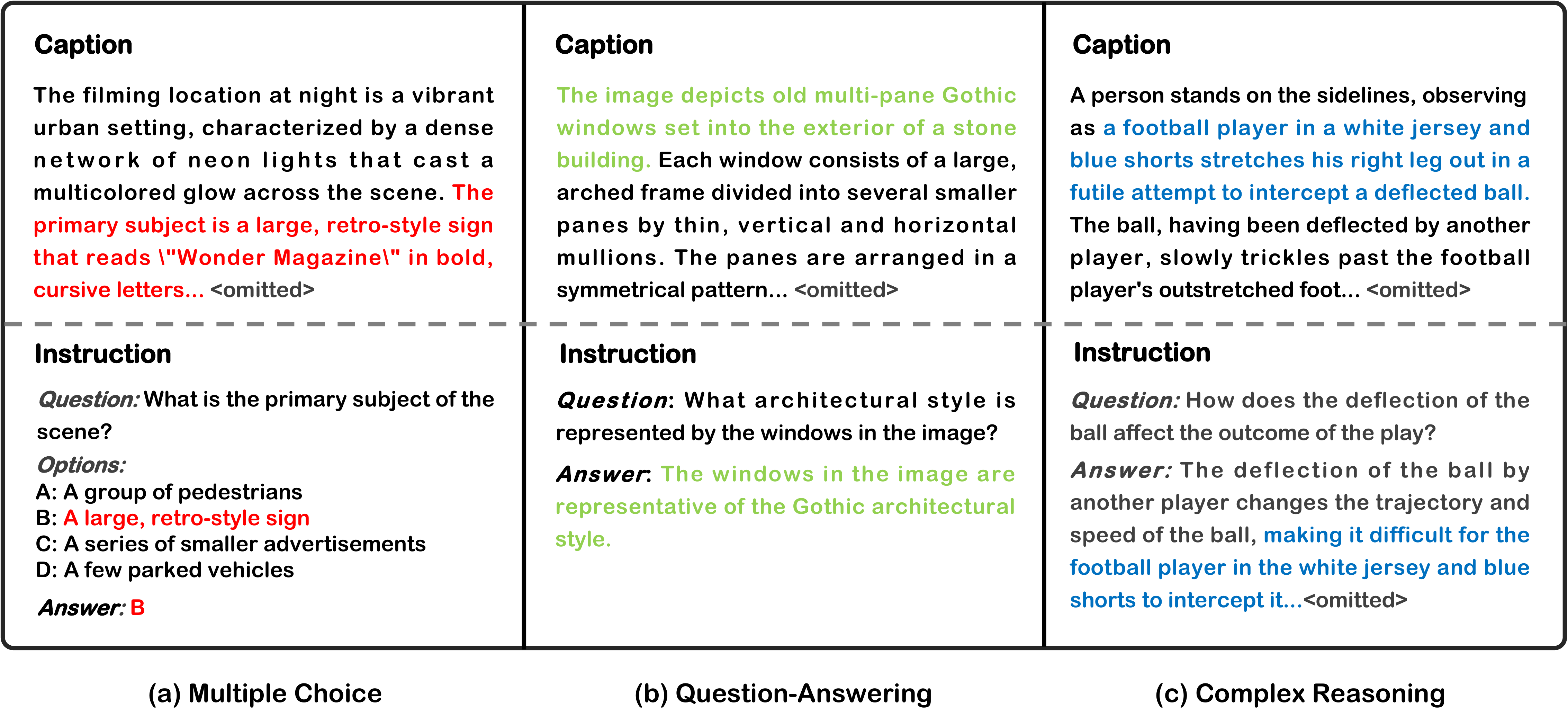}
  \caption{Data formats for the three instruction-tuning tasks. Each dialogue’s answer is derived from the textual content of diverse caption.}
  \label{fig:Instruction}
    
\end{figure*}

\subsection{Stage 1. Diverse Caption Data}

The first condition for text-only multimodal supervision is semantic visualizability: a textual description should contain sufficiently concrete visual information so that its corresponding visual equivalence set becomes less ambiguous. Therefore, given a sparse seed caption $s_i$, our goal is to generate a dense visual caption $c_i=\Phi(s_i)$ that specifies objects, attributes, spatial relations, background, lighting, and scene context. Formally, let $\mathcal{X}(s)$ denote the set of possible images that are semantically consistent with a textual description $s$. Semantic densification aims to reduce the uncertainty of the corresponding visual representation distribution, i.e.,
\begin{equation}
\mathrm{Var}_{x\sim\mathcal{X}(c_i)}[f_I(x)]<
\mathrm{Var}_{x\sim\mathcal{X}(s_i)}[f_I(x)].
\end{equation}
Thus, the dense caption $c_i$ can be regarded as describing a more specific imaginary visual scene, even though no real image is used.

Given the Unicorn-1.2M-Seed dataset $S=\{s_1,s_2,\ldots,s_N\}$, we use Qwen2.5-72B-Instruct \cite{qwen2025qwen25technicalreport} as the rewriting function $\Phi$ to enrich each seed caption $s_i$ with additional visual details, producing the dense visual caption $c_i=\Phi(s_i)$. The prompt is provided in the Appendix. This process produces a dense caption set $C=\{c_1,c_2,\ldots,c_N\}$ containing 1.2M samples. The resulting dense captions serve three roles in Unicorn: \ding{182} they provide caption-level supervision for pretraining; \ding{183} they serve as semantic sources for instruction synthesis in Stage 2; and \ding{184} after text encoding, they are transformed into synthetic image representations in Stage 3.

\subsection{Stage 2. Instruction-Tuning Data}

After semantic densification, each caption $c_i$ describes a concrete imaginary visual scene. However, the caption is insufficient for training instruction-following VLMs. Therefore, we further convert dense visual semantics into instruction-tuning data. This stage does not introduce images; it derives questions and answers directly from captions, ensuring that all supervision remains text-only and semantically consistent with $c_i$.

\begin{figure*}[h!]
  \centering
  \includegraphics[width=0.98\textwidth]{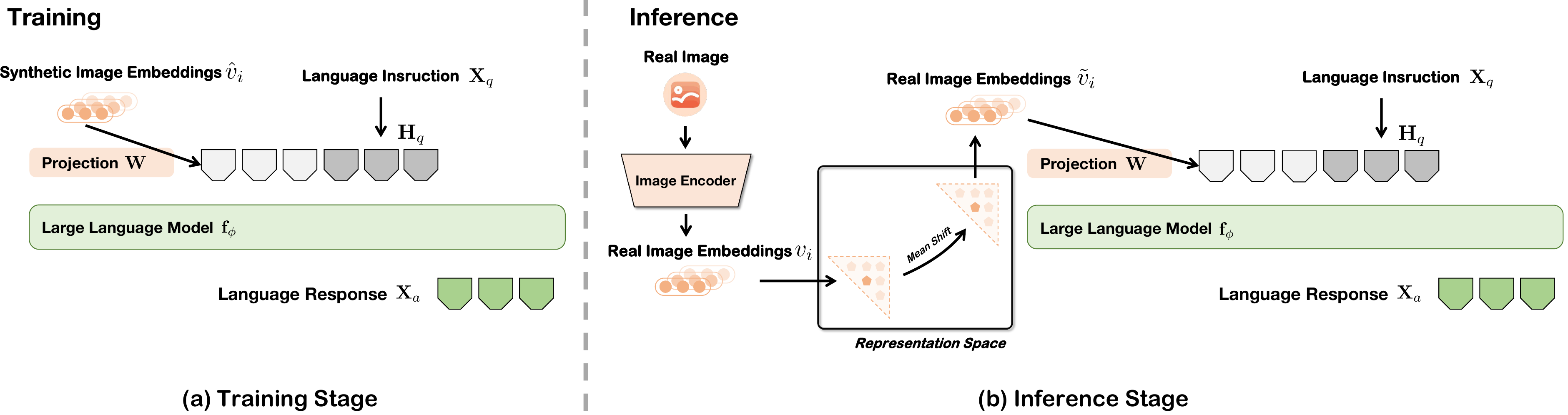}
  \caption{Training and inference processes of Unicorn-8B. Training aligns synthetic image representations with LLM embeddings, while inference adjusts real image embeddings as input. }
  \label{fig:process}
\end{figure*}

As shown in Fig.~\ref{fig:Instruction}, we design three instruction-tuning tasks. \ding{182} \textbf{Multiple Choice.} This task generates multiple-choice questions based on a given caption $c_i$. Each question contains four options, denoted as A, B, C, and D, with only one correct answer. \ding{183} \textbf{Question Answering.} This task generates open-ended questions and short answers based on the dense caption. \ding{184} \textbf{Complex Reasoning.} This task generates questions that require multi-step reasoning, hypothetical deduction, or commonsense inference grounded in the dense caption. For these tasks, we sample 171K, 199K, and 100K captions from the dense caption set $C$, respectively. For each selected caption $c_i$, we insert it into a task-specific template and use Qwen2.5-72B-Instruct \cite{qwen2025qwen25technicalreport} to generate an instruction-tuning sample $t_j$. Since all generated questions and answers are derived from the original dense caption, the resulting instruction data remain consistent with the underlying imaginary visual scene. This design mitigates hallucination compared with pipelines that synthesize image-text pairs without explicit semantic control. Finally, we synthesize a total of 471K high-quality instruction-tuning samples, denoted as $T=\{t_1,t_2,\ldots,t_M\}$.

\subsection{Stage 3. Representation Transfer}

The third stage implements the representation compatibility principle established in Sec.~\ref{sec2}. The empirical analysis shows that text embeddings and image embeddings are systematically shifted in the shared representation space, and that the dominant component of this shift is stable and transferable across semantic regions. Therefore, we synthesize visually compatible image representations by applying modality-gap-guided transfer in the shared embedding space.

Given the dense caption set $C=\{c_1,c_2,\ldots,c_N\}$, we encode each caption $c_i$ using the text encoder of LLM2CLIP~\citep{huang2026llm2clip}, obtaining a text representation $u_i\in\mathbb{R}^{d}$. The full set of dense-caption text representations is denoted as $U=\{u_1,u_2,\ldots,u_N\}$. We adopt LLM2CLIP because it can process long-form captions and complex semantic captions through its extended input window, making it suitable for encoding dense visual captions.

According to Sec.~\ref{sec2}, a calibrated text-to-visual transfer can be written as $u_i-\mu_T+\mu_V$, where $\mu_T$ and $\mu_V$ denote the centers of text and image representation distributions, respectively. However, in the text-only setting, real images are unavailable, and $\mu_V$ cannot be estimated from the synthetic data. Therefore, we use an image-free approximation that removes the dominant text-side bias:
\begin{equation}
\hat{v}_i=\mathrm{Norm}(u_i-\mu_U),
\end{equation}
where $\mu_U=|U|^{-1}\sum_i u_i$ is the mean of the text representations, and $\mathrm{Norm}(\cdot)$ denotes $\ell_2$ normalization. The resulting $\hat{v}_i$ is used as the synthetic image representation paired with caption $c_i$.

This operation should not be interpreted as reconstructing the real image representation of $c_i$. Instead, it constructs a visually compatible surrogate in the shared representation space. As shown in Sec.~\ref{sec2}, modality gap transfer improves text-image paired alignment across unseen semantic regions, suggesting that such a shared transfer rule can generalize to synthetic dense captions without paired images. Thus, Stage 3 converts text-only dense captions into representation-level visual supervision without generating pixels.

\subsection{Synthetic Dataset Construction}

After the three stages, Unicorn produces three components: dense visual captions $C$, instruction-tuning data $T$, and synthetic image representations $\hat{V}=\{\hat{v}_i\}_{i=1}^{N}$. These components are paired to construct two text-only multimodal datasets.

\noindent \textbf{Unicorn-1.2M.} For each dense caption $c_i$, we pair it with its synthetic image representation $\hat{v}_i$, forming a pretraining dataset: $\mathcal{D}_I=\{(\hat{v}_i,c_i)\}_{i=1}^{N}$. This dataset contains 1.2M synthetic multimodal pairs.

\noindent \textbf{Unicorn-471K-Instruction.} For each instruction-tuning sample $t_j$ generated from a dense caption $c_{a(j)}$, we pair it with the corresponding synthetic image representation $\hat{v}_{a(j)}$, forming an instruction-tuning dataset $\mathcal{D}_{II}=\{(\hat{v}_{a(j)},t_j)\}_{j=1}^{M}$, where $a(j)$ denotes the index of the dense caption from which instruction sample $t_j$ is generated. This dataset contains 471K synthetic instruction pairs.
\begin{table*}[t]
\centering
\setlength{\tabcolsep}{2pt}
\scriptsize
\caption{Ablation study on modality representation transfer. We compare directly using text embeddings, applying C$^3$ Align, and the proposed transfer in \method{}.} 
\label{tab1}
\renewcommand{\arraystretch}{1.2}
\resizebox{\textwidth}{!}{%
\begin{tabular}{lccccccccccccc}
\toprule
\multirow{2}{*}{\textbf{Method}}
& \multirow{2}{*}{\textbf{Text-only}}
& \multicolumn{4}{c}{\textbf{General}}
& \multicolumn{4}{c}{\textbf{Reasoning}}
& \multicolumn{3}{c}{\textbf{Hallucination}}
& \multirow{2}{*}{\textbf{Avg. $\uparrow$}} \\
\cmidrule(lr){3-6} \cmidrule(lr){7-10} \cmidrule(lr){11-13}
&& \texttt{MME} & \texttt{MMStar} & \texttt{SQA} & \texttt{RealWorldQA}
& \texttt{MMMU} & \texttt{MMMU-P} & \texttt{VisuLogic} & \texttt{LogicVista}
& \texttt{CRPE} & \texttt{POPE} & \texttt{HallBench} & \\
\midrule
w/o. Align & \cmark 
& 57.46 & \underline{33.87} & \underline{66.02} & \underline{39.18}
& 33.41 & \underline{31.26} & \underline{24.10} & 26.44
& \underline{39.07} & \underline{62.37} & \underline{40.12} & \underline{41.21} \\
C$^3$ Align & \cmark 
& \underline{58.02} & 32.13 & 65.74 & 38.66
& \underline{33.95} & 30.72 & 23.90 & \underline{26.97}
& 38.64 & 61.31 & 39.58 & 40.87 \\
\rowcolor{graybg} \method{} & \cmark 
& \textbf{60.24} & \textbf{35.13} & \textbf{68.81} & \textbf{42.35}
& \textbf{36.87} & \textbf{34.05} & \textbf{26.80} & \textbf{29.53}
& \textbf{42.32} & \textbf{64.21} & \textbf{43.01} & \textbf{43.94} \\
\hdashline
\bottomrule
\end{tabular}%
}
\end{table*}

\begin{table*}[t]
\centering
\setlength{\tabcolsep}{2pt}
\scriptsize
\caption{
Scale-matched comparison between paired image-text pretraining and text-only pretraining. For each scale, Unicorn is matched to the pretraining data size of the corresponding image-text baseline, while all baselines use their original pretraining data. All methods use the same 417K SFT data.
}
\label{tab2}
\renewcommand{\arraystretch}{1.18}
\resizebox{\textwidth}{!}{%
\begin{tabular}{lcccccccccccccc}
\toprule
\multirow{2}{*}{\textbf{Method}}
& \multirow{2}{*}{\textbf{Text-only}}
& \multirow{2}{*}{\textbf{Pretrain Scale}}
& \multicolumn{4}{c}{\textbf{General}}
& \multicolumn{4}{c}{\textbf{Reasoning}}
& \multicolumn{3}{c}{\textbf{Hallucination}}
& \multirow{2}{*}{\textbf{Avg. $\uparrow$}} \\
\cmidrule(lr){4-7} \cmidrule(lr){8-11} \cmidrule(lr){12-14}
&&& \texttt{MME} & \texttt{MMStar} & \texttt{SQA} & \texttt{RealWorldQA}
& \texttt{MMMU} & \texttt{MMMU-P} & \texttt{VisuLogic} & \texttt{LogicVista}
& \texttt{CRPE} & \texttt{POPE} & \texttt{HallBench} & \\
\midrule

LLaVA-1.5 & \xmark & 558K
& 54.86 & 33.27 & 64.91 & 38.73
& 33.16 & 29.88 & 22.90 & 25.41
& 37.66 & 60.73 & 38.84 & 40.03 \\

\method{}$^*$ & \cmark & 558K
& 56.71 & 34.20 & 65.12 & 39.83
& 34.23 & 30.27 & 24.30 & 26.81
& 39.22 & 62.14 & 40.38 & 41.20 \\

\hdashline

Bunny & \xmark & 1.2M
& \underline{58.37} & \underline{34.73} & \underline{67.28} & \underline{41.08}
& 34.62 & \underline{32.41} & \underline{25.30} & \underline{27.86}
& \underline{40.51} & \underline{63.12} & \underline{41.26} & \underline{42.41} \\

\rowcolor{graybg} \method{} & \cmark & 1.2M
& \textbf{60.24} & \textbf{35.13} & \textbf{68.81} & \textbf{42.35}
& \textbf{36.87} & \textbf{34.05} & \textbf{26.80} & \textbf{29.53}
& \textbf{42.32} & \textbf{64.21} & \textbf{43.01}
& \textbf{43.94} \\

\bottomrule
\end{tabular}%
}
\end{table*}

\section{Unicorn-8B}

Based on the two synthetic datasets, we train a VLM referred to as Unicorn-8B. This section presents the Unicorn-8B's architecture and details of its training and inference processes. Fig. \ref{fig:process} shows the relevant details.

\subsection{Architecture} 

Unicorn-8B is composed of two components: the projector $W$ and the backbone LLM $\boldsymbol{f_\varphi}$. For the projector $W$, we adopt a multilayer perceptron (MLP) architecture equipped with GELU activation functions, following the design in LLaVA \cite{liu2024improved}. This component serves as a modality adapter, effectively bridging the input representations to the latent space of the LLM. We utilize LLaMA3-8B-Instruct as the backbone LLM.

\subsection{Training \& Inference}


\textbf{Training.} In the pretraining stage, synthetic image representations $\hat{v}_i$ are aligned with pre-trained LLM representation $H_{q}$ through the projection matrix $W$. During this phase, only $W$ is updated, while the weights of the LLM $\boldsymbol{f_\varphi}$ remain frozen. Subsequently, in the instruction-tuning stage, the Unicorn-471K-Instruction is employed to update both $W$ and the LLM $\boldsymbol{f_\varphi}$ weights, thereby enhancing the model’s instruction-following capability.

\noindent \textbf{Inference.} In the inference stage, real images from the test set are decoded by the image encoder to get real image representations. 
To mitigate the modality gap, the global mean of the test image representation set $\mathbb{E}[V]$ is subtracted from each input test image representation $v_i \in \mathbb{R}^h$, resulting in a transformed representation $\tilde{v}_i = v_i - \mathbb{E}[V]$. 
\section{Experiments}

\begin{table*}[t]
\centering
\setlength{\tabcolsep}{2pt}
\scriptsize
\caption{
Ablation study on semantic densification. 
The variant w/o Semantic Densification directly uses raw caption seeds for pretraining, while \method{} uses dense visual captions.
}
\label{tab3}
\renewcommand{\arraystretch}{1.18}
\resizebox{\textwidth}{!}{%
\begin{tabular}{lcccccccccccc}
\toprule
\multirow{2}{*}{\textbf{Method}}
& \multicolumn{4}{c}{\textbf{General}}
& \multicolumn{4}{c}{\textbf{Reasoning}}
& \multicolumn{3}{c}{\textbf{Hallucination}}
& \multirow{2}{*}{\textbf{Avg. $\uparrow$}} \\
\cmidrule(lr){2-5} \cmidrule(lr){6-9} \cmidrule(lr){10-12}
& \texttt{MME} & \texttt{MMStar} & \texttt{SQA} & \texttt{RealWorldQA}
& \texttt{MMMU} & \texttt{MMMU-P} & \texttt{VisuLogic} & \texttt{LogicVista}
& \texttt{CRPE} & \texttt{POPE} & \texttt{HallBench} & \\
\midrule
w/o. Dense
& 57.36 & 33.87 & 66.14 & 39.72
& 33.96 & 31.42 & 24.30 & 26.88
& 39.51 & 62.43 & 40.28 & 41.44 \\

\rowcolor{graybg} \method{}
& \textbf{60.24} & \textbf{35.13} & \textbf{68.81} & \textbf{42.35}
& \textbf{36.87} & \textbf{34.05} & \textbf{26.80} & \textbf{29.53}
& \textbf{42.32} & \textbf{64.21} & \textbf{43.01}
& \textbf{43.94} \\
\bottomrule
\end{tabular}%
}
\end{table*}

\begin{table*}[t]
\centering
\setlength{\tabcolsep}{2pt}
\scriptsize
\caption{
Ablation study on seed diversity. Open-domain uses open-domain captions, while \method{} uses both open-domain captions and domain-specific knowledge captions.
}
\label{tab4}
\renewcommand{\arraystretch}{1.18}
\resizebox{\textwidth}{!}{%
\begin{tabular}{lcccccccccccc}
\toprule
\multirow{2}{*}{\textbf{Method}}
& \multicolumn{4}{c}{\textbf{General}}
& \multicolumn{4}{c}{\textbf{Reasoning}}
& \multicolumn{3}{c}{\textbf{Hallucination}}
& \multirow{2}{*}{\textbf{Avg. $\uparrow$}} \\
\cmidrule(lr){2-5} \cmidrule(lr){6-9} \cmidrule(lr){10-12}
& \texttt{MME} & \texttt{MMStar} & \texttt{SQA} & \texttt{RealWorldQA}
& \texttt{MMMU} & \texttt{MMMU-P} & \texttt{VisuLogic} & \texttt{LogicVista}
& \texttt{CRPE} & \texttt{POPE} & \texttt{HallBench} & \\
\midrule
Open-domain
& 58.86 & 34.27 & 67.56 & 40.91
& 34.95 & 32.18 & 25.10 & 27.42
& 40.13 & 63.48 & 41.08 & 42.36 \\

\rowcolor{graybg} \method{}
& \textbf{60.24} & \textbf{35.13} & \textbf{68.81} & \textbf{42.35}
& \textbf{36.87} & \textbf{34.05} & \textbf{26.80} & \textbf{29.53}
& \textbf{42.32} & \textbf{64.21} & \textbf{43.01}
& \textbf{43.94} \\
\bottomrule
\end{tabular}%
}
\end{table*}

We evaluate \method{} on three benchmark groups: general visual understanding, multimodal reasoning, and hallucination. Detailed experimental settings are provided in Appendix~\ref{App:E}.

\noindent \paragraph{Modality Representation Transfer.} Table~\ref{tab1} compares alignment strategies for constructing synthetic image representations. Directly using text embeddings without alignment achieves an average score of $41.21$, indicating that text embeddings preserve useful semantics but remain biased toward the text-side distribution. C$^3$ \cite{zhang2024connect} Align obtains a lower average score of $40.87$, despite improving a few individual metrics. This suggests that adding stochastic or Gaussian-style perturbations to model residual uncertainty is insufficient: the residual term $\epsilon_i$ in the modality-gap model is not the dominant factor to correct before the global bias is removed, and isotropic noise may even disturb semantic consistency. In contrast, Unicorn explicitly compensates for the structured modality shift and achieves the best average score of $43.94$, improving over w/o Align by $+2.73$. The gains are consistent across general, reasoning, and hallucination benchmarks, supporting the theoretical claim that text embeddings must be transferred toward the visual-compatible distribution rather than used directly or perturbed with generic noise.

\noindent \paragraph{Image-Text Training Baselines.} Table~\ref{tab2} compares Unicorn with paired image-text pretraining baselines. To ensure a fair comparison, all methods are matched in data scale. Under the same training data scale setting, Unicorn$^*$ achieves an average score of $41.20$, outperforming LLaVA-1.5 by $+1.17$. Unicorn further reaches $43.94$, surpassing Bunny by $+1.53$. The gains are especially clear on reasoning-oriented benchmarks such as MMMU, MMMU-Pro, VisuLogic, and LogicVista. We attribute this improvement to the dense captions used in Unicorn: compared with conventional image-text captions, they provide more explicit descriptions of objects, attributes, spatial relations, and scene context, thereby offering richer semantic supervision for VLM training. These results indicate that, with semantic densification and modality-gap-guided transfer, text-only representation-level supervision can serve as an effective alternative to standard paired image-text supervision.

\noindent \paragraph{Ablation on Semantic Densification.} Table~\ref{tab3} evaluates the role of semantic densification. Removing this stage and using raw caption seeds reduces the average score from $43.94$ to $41.44$. This drop supports the semantic visualizability assumption in Sec.~\ref{sec2}: sparse captions correspond to a broad visual equivalence set $\mathcal{X}(s)$, making the induced visual representation distribution ambiguous. Dense captions reduce this uncertainty by specifying objects, attributes, relations, background, and scene context. The performance degradation is particularly evident on reasoning and hallucination-related benchmarks, indicating that semantic densification does not merely increase text length but provides more determinate visual semantics for constructing stable synthetic image representations.

\paragraph{Ablation on Seed Diversity.} Table~\ref{tab4} studies the effect of seed diversity. Using only open-domain caption seeds yields an average score of $42.36$, while the full Unicorn setting reaches $43.94$. The improvement is larger on reasoning benchmarks, where domain-specific knowledge seeds provide fine-grained concepts that are less covered by open-domain captions. This result is consistent with the observation in Sec.~\ref{sec2} that the dominant modality gap is shared across semantic regions: a unified transfer rule can be applied to diverse textual sources, but the semantic coverage of those sources determines the richness of the supervision. Therefore, combining open-domain and domain-specific seeds improves both coverage and precision.


\noindent \paragraph{Cost Analysis.} We provide cost analysis in Appendix~\ref{App:D}. Since Unicorn operates representation spaces, it avoids image collection, image storage, and image-conditioned annotation, making it practical for scalable VLM data construction.

\noindent \paragraph{Scale Analysis.} Additional analyses on data scales are provided in Appendix~\ref{App:E}. The results show that Unicorn benefits from increased text-only supervision, indicating that the proposed synthesis pipeline can effectively scale with data size.


\section{Conclusion}



In this paper, we present \method{}, a text-only data synthesis framework for VLM training. By analyzing the modality gap in shared representation space, we show that text embeddings can be transferred into visually compatible representations through a stable and transferable modality gap structure. Guided by this finding, \method{} constructs VLM training data through a three-stage pipeline without using real images. Experiments show that \method{} achieves competitive performance against paired image-text pretraining baselines. These results demonstrate that text-only data can be a scalable and effective source of multimodal supervision, providing an alternative to conventional image-dependent data construction. We hope \method{} inspires future works on scalable synthetic data.


\newpage

\section*{Limitations}

This work mainly focuses on VLM training in the image-text setting and has not yet been extended to broader multimodal inputs such as video, audio, or 3D data. Future work may explore whether similar text-only synthesis principles can be transferred to other modality forms, further broadening the applicability of the proposed framework.

\bibliography{custom}

\newpage
\appendix
\onecolumn


\section{Theoretical Justification}
\label{app:theoretical_justification}

This section provides a theoretical justification for the modality-gap-guided representation transfer used in Unicorn. The analysis complements the empirical principles in Section~\ref{sec2} and formalizes why a shared gap-transfer rule can convert text representations into visually compatible representations in the shared embedding space.

\subsection{Global Modality Bias as the Optimal Translation}

Let $u_i=f_T(y_i)$ and $v_i=f_I(x_i)$ denote the text and image representations of a paired sample. We model their relationship as
\begin{equation}
v_i = u_i + c + \epsilon_i, \qquad \mathbb{E}[\epsilon_i]=0,
\end{equation}
where $c$ denotes the modality-level bias and $\epsilon_i$ denotes the sample-level residual discrepancy. Under the squared-error criterion, the optimal global transfer vector is
\begin{equation}
c^*=\arg\min_c \mathbb{E}\left[\|u_i+c-v_i\|_2^2\right].
\end{equation}
Taking the derivative with respect to $c$ gives
\begin{equation}
\mathbb{E}[u_i+c^*-v_i]=0,
\end{equation}
and therefore
\begin{equation}
c^*=\mathbb{E}[v_i-u_i].
\end{equation}
Thus, the empirical global gap $\bar{\Delta}=N^{-1}\sum_i(v_i-u_i)$ is the finite-sample estimator of the optimal translation. This shows that the global modality gap is not an ad-hoc correction, but the optimal constant transfer vector under the mean-squared alignment objective.

The corresponding transferred representation is $\hat{v}_i=u_i+\bar{\Delta}$. Its error can be written as
\begin{equation}
\hat{v}_i-v_i=(\bar{\Delta}-c)-\epsilon_i.
\end{equation}
When $\bar{\Delta}$ consistently estimates $c$, the remaining error is dominated by the residual term $\epsilon_i$. Therefore, global transfer should be understood as improving visual compatibility rather than exactly reconstructing each image representation.

\subsection{Consistency of the Empirical Gap Estimator}

Assume that $\Delta_i=v_i-u_i$ are independent samples with finite covariance $\Sigma_\Delta$, and that $\mathbb{E}[\Delta_i]=c$. The empirical estimator is
\begin{equation}
\bar{\Delta}=\frac{1}{N}\sum_{i=1}^{N}\Delta_i.
\end{equation}
Then
\begin{equation}
\mathbb{E}[\bar{\Delta}]=c,
\qquad
\mathrm{Cov}(\bar{\Delta})=\frac{1}{N}\Sigma_\Delta.
\end{equation}
Hence, the estimation error decreases as $O(N^{-1/2})$. Equivalently,
\begin{equation}
\mathbb{E}\left[\|\bar{\Delta}-c\|_2^2\right]
=
\frac{1}{N}\mathrm{Tr}(\Sigma_\Delta).
\end{equation}
This explains why estimating the modality gap from a large paired representation set is stable. Although the residual term $\epsilon_i$ may remain large at the instance level, the global component can still be estimated reliably from sufficiently many samples.

\subsection{Why Cross-Region Transfer Improves Alignment}

We next analyze why a gap estimated from one semantic region can improve alignment in another region. Let $\mathcal{R}_s$ and $\mathcal{R}_t$ denote source and target semantic regions. Suppose the local gap in each region can be written as
\begin{equation}
c_s = c + r_s, \qquad c_t = c + r_t,
\end{equation}
where $c$ is the shared modality-level bias and $r_s,r_t$ are region-specific residual biases. For a target sample, assume
\begin{equation}
v_i = u_i + c_t + \epsilon_i, \qquad \mathbb{E}[\epsilon_i]=0.
\end{equation}
Without transfer, the expected squared alignment error is
\begin{equation}
\mathcal{E}_{0}
=
\mathbb{E}\left[\|u_i-v_i\|_2^2\right]
=
\|c_t\|_2^2+\mathbb{E}\left[\|\epsilon_i\|_2^2\right].
\end{equation}
If we estimate a transfer vector from the source region and apply it to the target region, the transferred error becomes
\begin{equation}
\mathcal{E}_{s\rightarrow t}
=
\mathbb{E}\left[\|u_i+c_s-v_i\|_2^2\right]
=
\|c_s-c_t\|_2^2+\mathbb{E}\left[\|\epsilon_i\|_2^2\right].
\end{equation}
Therefore, source-to-target transfer improves the expected squared error whenever
\begin{equation}
\|c_s-c_t\|_2^2 < \|c_t\|_2^2.
\end{equation}
Equivalently,
\begin{equation}
2\langle c_s,c_t\rangle > \|c_s\|_2^2.
\end{equation}
This condition holds when the source and target gap directions are sufficiently aligned and the region-specific residual biases are small relative to the shared global bias. This directly corresponds to the empirical observations in Section~\ref{sec2}: local gaps remain positively aligned with the global gap, and a source-estimated gap consistently improves target-region cosine similarity.

For normalized representations, reducing squared distance is closely related to improving cosine similarity. If $a$ and $b$ are $\ell_2$-normalized, then
\begin{equation}
\|a-b\|_2^2 = 2-2\cos(a,b).
\end{equation}
Thus, a reduction in alignment error corresponds to an increase in paired cosine similarity. This explains why the cross-cluster transfer experiment can be interpreted as evidence for transferable visual compatibility.

\subsection{Semantic Densification Reduces Visual Ambiguity}

The first stage of Unicorn expands a sparse caption $s_i$ into a dense visual caption $c_i=\Phi(s_i)$. This operation is motivated by the fact that sparse captions are compatible with many visually different images. Let $\mathcal{X}(s)$ denote the set of images semantically consistent with a textual description $s$, and let $Z=f_I(x)$ be the corresponding image representation. A dense caption can be viewed as a refinement of the sparse caption because it introduces additional visual constraints, such as objects, attributes, relations, background, and scene context.

Formally, let $S$ denote a sparse description and $C$ denote its densified description. Since $C$ contains additional information conditioned on $S$, the law of total variance gives
\begin{equation}
\mathrm{Var}(Z\mid S)
=
\mathbb{E}\left[\mathrm{Var}(Z\mid C)\mid S\right]
+
\mathrm{Var}\left(\mathbb{E}[Z\mid C]\mid S\right).
\end{equation}
Since the second term is non-negative, we obtain
\begin{equation}
\mathbb{E}\left[\mathrm{Var}(Z\mid C)\mid S\right]
\leq
\mathrm{Var}(Z\mid S).
\end{equation}
This shows that conditioning on a denser visual description reduces the expected uncertainty of the corresponding visual representation. Therefore, semantic densification does not merely increase the length of captions; it reduces visual ambiguity and makes the text representation closer to a specific visual-compatible region in the shared embedding space.

\subsection{Image-Free Approximation in Unicorn}

The ideal text-to-visual transfer based on paired statistics can be written as
\begin{equation}
\hat{v}_i = \mathrm{Norm}(u_i-\mu_T+\mu_V),
\end{equation}
where $\mu_T$ and $\mu_V$ are the centers of text and image representation distributions. However, in the text-only synthesis setting, real images are unavailable, and therefore $\mu_V$ cannot be estimated from the synthetic data. This makes the exact paired mean-shift transfer unidentifiable without additional assumptions.

Unicorn therefore adopts an image-free approximation:
\begin{equation}
\hat{v}_i=\mathrm{Norm}(u_i-\mu_U),
\end{equation}
where $\mu_U=|U|^{-1}\sum_i u_i$ is the center of the dense-caption text representations. This operation removes the dominant text-side mean component and normalizes the resulting representation. It should not be interpreted as an unbiased estimator of the real image embedding. Instead, it constructs a visually compatible surrogate by reducing text-side distributional bias and placing the synthetic representation in a normalized shared embedding space.

Combined with the empirical transferability of the modality gap, this approximation provides a practical image-free mechanism for constructing representation-level visual supervision. The resulting synthetic image representations are therefore best understood as visual-compatible training signals, rather than exact substitutes for real image features.


\section{Additional Analysis}
\label{App:B}

\begin{figure*}[t]
    \centering
    \includegraphics[width=0.95\textwidth]{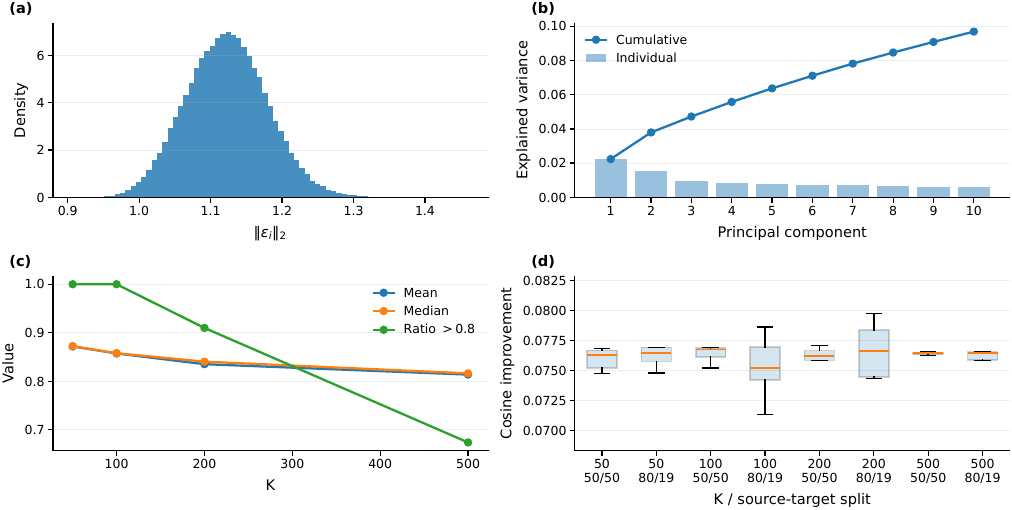}
    \caption{Additional analysis of the modality gap. (a) Residual norm distribution after removing the global gap. (b) PCA explained variance of pairwise gap vectors. (c) Cluster-level statistics across different semantic granularities. (d) Cross-cluster transfer improvement on disjoint target clusters.}
    \label{fig_data2}
\end{figure*}

This section provides additional analyses for the modality-gap principles discussed in Sec.~\ref{sec2}.

\paragraph{Residual structure after removing the global bias.}
In Sec.~\ref{sec2}, we decompose the pairwise image-text gap as $\Delta_i=\bar{\Delta}+\epsilon_i$, where $\bar{\Delta}$ denotes the global modality bias and $\epsilon_i$ denotes the residual term. To further examine the residual structure, we analyze the norm distribution of $\epsilon_i=\Delta_i-\bar{\Delta}$. As shown in Fig.~\ref{fig_data2}(a), the residual norm is mainly concentrated between $1.05$ and $1.20$, indicating that substantial instance-level discrepancy remains after removing the global gap. Meanwhile, the residual mean norm is close to zero, suggesting that the residuals are approximately centered after global-bias removal. This supports our modeling choice: the modality gap should not be treated as a deterministic translation alone, but as a combination of a stable global bias and sample-level residual variation. Therefore, the purpose of representation transfer is not to reconstruct the exact image embedding for each sample, but to construct a visually compatible approximation in the shared representation space.

\paragraph{Dimensionality of pairwise gap variations.}
We further perform PCA on the pairwise gap vectors $\Delta_i$. As shown in Fig.~\ref{fig_data2}(b), the first principal component explains only about $2.2\%$ of the variance, and the top ten components explain about $9.7\%$ in total. This indicates that the residual variation of the modality gap is not concentrated in a single dominant direction. In other words, although the global gap $\bar{\Delta}$ provides a stable and useful modality-level correction, the full pairwise discrepancy contains richer sample-level variations. This observation is consistent with the residual analysis above and explains why a single global vector improves visual compatibility but cannot precisely recover real image representations.

\paragraph{Cluster-level stability across semantic granularities.}
In the main text, we show that local gaps estimated from unsupervised semantic clusters remain positively aligned with the global gap. Fig.~\ref{fig_data2}(c) provides a complementary summary across different values of $K$. The mean and median similarities decrease mildly as $K$ increases, while remaining consistently positive. This trend is expected because larger $K$ produces finer clusters with fewer samples per cluster, making each local gap estimate noisier. The ratio of clusters with similarity larger than $0.8$ also decreases as semantic regions become finer, indicating that semantic-specific residuals become more visible at fine granularities. Nevertheless, the local gaps remain substantially aligned with the global gap across all settings, supporting the conclusion that the dominant modality bias is shared across semantic regions rather than being fully determined by specific categories.

\paragraph{Robustness of cross-cluster transfer.}
Finally, we provide the distribution of cosine improvements in the cross-cluster transfer experiment. As shown in Fig.~\ref{fig_data2}(d), the source-estimated gap consistently improves paired cosine similarity on disjoint target clusters across different clustering granularities and source/target split ratios. The improvements are concentrated around $+0.075$, showing that the gain is stable rather than caused by a particular split. Since the target clusters are not used to estimate the transfer vector, this result further supports the cross-region transferability of the modality gap. It also justifies applying a shared gap-transfer rule to synthetic dense captions without paired images.


\section{Data Synthesis Prompts}

This section provides the prompts used in the text-only data synthesis pipeline. The prompts are used for two purposes. First, the detailed caption prompt expands sparse caption seeds into dense visual descriptions, making the textual input more visually specific and less ambiguous. Second, the instruction-synthesis prompts convert dense visual captions into three types of instruction-following data: multiple choice, visual question answering, and complex reasoning. All prompts operate only on textual inputs and do not require real or synthetic images.

\begin{tcolorbox}[promptbox,title=Detailed Caption]
\small
Please expand the given image caption to include observable, objective, and reasonable details, avoiding any literary rhetoric, subjective judgments, or emotional tones. Please focus especially on depicting the detailed features of the primary subject, ensuring a clear and specific caption of its appearance and structure, while making all details as comprehensive and concrete as possible. Ensure the caption remains natural and all content is presented in a single paragraph. The image caption is as follows: \texttt{<seed>}
\end{tcolorbox}

\begin{tcolorbox}[promptbox,title=Task 1: Multiple Choice]
\small
You MUST ONLY output valid JSON. Generate nothing but JSON! The JSON must have the structure:

\vspace{2pt}
\begin{lstlisting}[style=promptjson]
{
  "id": " ",
  "QA Pairs": [
    {
      "Question": " ",
      "Options": {
        "A": " ",
        "B": " ",
        "C": " ",
        "D": " "
      },
      "Answer": " "
    },
    ...
  ]
}
\end{lstlisting}

\vspace{2pt}
Please generate exactly 6 to 8 vision question-answer pairs related to the image caption. Each question should have four options (A, B, C, D), with exactly one correct answer. The questions should be diverse in type and complexity, ensuring a range of reasoning skills (e.g., ``when,'' ``why,'' ``what,'' ``how,'' ``does/do,'' ``is/are,'' ``where,'' ``who,'' ``which one,'' etc.). Each question should require integrating and analyzing the overall information from the image caption, covering various aspects of the image caption content.

\vspace{3pt}
The image caption is as follows: \texttt{<caption>}
\end{tcolorbox}

\begin{tcolorbox}[promptbox,title=Task 2: Visual Question Answering]
\small
You MUST ONLY output valid JSON. Generate nothing but JSON! The JSON must have the structure:

\vspace{2pt}
\begin{lstlisting}[style=promptjson]
{
  "id": " ",
  "QA Pairs": [
    {
      "Question": " ",
      "Answer": " "
    },
    ...
  ]
}
\end{lstlisting}

\vspace{2pt}
Please generate exactly 6 to 8 vision question-answer pairs based on the image caption. Incorporate a variety of question types---such as ``what,'' ``why,'' ``how,'' ``does/do,'' ``is/are,'' ``where,'' ``who,'' and ``which,''---to explore different aspects and details of the image. Ensure responses incorporate diverse and authentic language, steering clear of verbatim phrases from the image caption. Address diverse elements of the image by examining chronological sequences, spatial relationships, causal links, and inferential insights to showcase a comprehensive understanding.

\vspace{3pt}
The image caption is as follows: \texttt{<caption>}
\end{tcolorbox}

\begin{tcolorbox}[promptbox,title=Task 3: Complex Reasoning]
\small
You MUST ONLY output valid JSON. Generate nothing but JSON! The JSON must have the structure:

\vspace{2pt}
\begin{lstlisting}[style=promptjson]
{
  "id": " ",
  "QA Pairs": [
    {
      "Question": " ",
      "Answer": " "
    },
    ...
  ]
}
\end{lstlisting}

\vspace{2pt}
Design a collection of intricate reasoning challenges based on the image caption. Each problem should present a unique scenario requiring thorough analysis, such as multi-step deductions, hypothetical reasoning, complex problem-solving, or cause-and-effect evaluations. Provide comprehensive, step-by-step solutions for each problem, clearly outlining the logical progression and inference methods used. Ensure the challenges vary in format and complexity, promoting deep understanding and stimulating intellectual engagement.

\vspace{3pt}
The image caption is as follows: \texttt{<caption>}
\end{tcolorbox}

\section{Synthetic Data Quality Evaluation} \label{App:D}

In this section, we evaluate the quality and scalability of the synthetic data produced by Unicorn. Since Unicorn is designed as an image-free data synthesis pipeline, our analysis focuses on three aspects that are directly relevant to large-scale text-only multimodal supervision: cost efficiency, length distribution, and lexical diversity.

        

        

\begin{table}[t]
    \centering
    \caption{Construction cost of \method{}-1.2M.}
    \label{tab:cost}
    \setlength{\tabcolsep}{6pt}
    \resizebox{0.42\textwidth}{!}{%
    \begin{tabular}{lccc}
        \toprule[1pt]
        \textbf{Dataset} & \textbf{API$/10^{-3}\$$} & \textbf{Time / Day} & \textbf{Storage / GB} \\
        \midrule
        \method{}-1.2M & 0.30 & 12 & 4 \\
        \bottomrule[1pt]
    \end{tabular}
    }
\end{table}





\begin{table}[t]
\small
\centering
\renewcommand\arraystretch{1.2}
\setlength{\tabcolsep}{6pt}
\caption{Diversity statistics of \method{}-1.2M.}
\label{tab:diversity}
\resizebox{0.42\textwidth}{!}{%
\begin{tabular}{lccc}
\toprule[1pt]
\textbf{Dataset} & \textbf{Scale} & \textbf{TTR $\uparrow$} & \textbf{Entropy $\uparrow$} \\
\midrule
\method{}-1.2M & 1,243,195 & 0.68 & 6.07 \\
\bottomrule[1pt]
\end{tabular}
}
\end{table}

\textbf{Cost Analysis.}
Unicorn avoids the collection, generation, and storage of real or synthetic images. As a result, the major cost of data construction comes from text-only LLM rewriting, instruction synthesis, and representation extraction. As shown in Table~\ref{tab:cost}, constructing Unicorn-1.2M requires an API cost of approximately \$0.0003 per sample, 12 days of generation time, and 4 GB of storage. These numbers indicate that the image-free design substantially reduces the practical overhead of large-scale multimodal data construction. In particular, because Unicorn stores textual data and synthetic representations rather than image files, its storage requirement remains lightweight even at the million-sample scale. This makes the proposed pipeline feasible for scalable VLM pretraining without relying on expensive image acquisition or image-conditioned annotation.

\textbf{Length Distribution.}
We further analyze the length distribution of Unicorn-1.2M to examine whether the generated dense captions provide sufficiently rich visual semantics. The caption lengths form a concentrated and smooth distribution rather than being dominated by very short or excessively long samples. The average length is around 129 words, with a minimum length of 51 words and a maximum length of 483 words. This suggests that the semantic densification stage effectively expands sparse caption seeds into detailed visual descriptions while maintaining a controlled length range. Therefore, Unicorn-1.2M provides richer supervision than short caption templates, yet avoids overly verbose descriptions that may introduce irrelevant textual noise.

\textbf{Diversity Analysis.}
To assess lexical diversity, we compute two standard statistics: Type-Token Ratio (TTR) and token-level entropy. As shown in Table~\ref{tab:diversity}, Unicorn-1.2M achieves a TTR of 0.68 and an entropy of 6.07. These results indicate that the generated captions contain a diverse vocabulary and are not limited to repetitive template-like expressions. This diversity mainly comes from two sources. First, the open-domain caption seeds cover a broad range of everyday visual scenarios. Second, the domain-specific knowledge seeds introduce fine-grained concepts from specialized domains. Together, these two sources allow Unicorn-1.2M to provide both broad semantic coverage and detailed visual concepts, which are important for constructing scalable text-only multimodal supervision.


\begin{table*}[t]
\centering
\setlength{\tabcolsep}{2pt}
\scriptsize
\caption{Effect of pretraining data scale. We vary the proportion of \method{}-1.2M used for modality-alignment pretraining while keeping the full 417K instruction-tuning data fixed.}
\label{tab6}
\renewcommand{\arraystretch}{1.18}
\resizebox{0.95\textwidth}{!}{%
\begin{tabular}{lcccccccccccc}
\toprule
\multirow{2}{*}{\textbf{Pretrain Ratio}}
& \multicolumn{4}{c}{\textbf{General}}
& \multicolumn{4}{c}{\textbf{Reasoning}}
& \multicolumn{3}{c}{\textbf{Hallucination}}
& \multirow{2}{*}{\textbf{Avg. $\uparrow$}} \\
\cmidrule(lr){2-5} \cmidrule(lr){6-9} \cmidrule(lr){10-12}
& \texttt{MME} & \texttt{MMStar} & \texttt{SQA} & \texttt{RealWorldQA}
& \texttt{MMMU} & \texttt{MMMU-P} & \texttt{VisuLogic} & \texttt{LogicVista}
& \texttt{CRPE} & \texttt{POPE} & \texttt{HallBench} & \\
\midrule
25\%
& 55.96 & 33.13 & 65.18 & 38.92
& 32.84 & 29.63 & 23.40 & 25.76
& 38.04 & 60.83 & 38.92 & 40.24 \\

50\%
& 58.34 & 34.27 & 67.42 & 40.91
& 35.03 & 32.01 & 25.10 & 27.81
& 40.62 & 62.94 & 41.18 & 42.33 \\

75\%
& \underline{59.43} & \underline{34.87} & \underline{68.17} & \underline{41.78}
& \underline{36.12} & \underline{33.18} & \underline{26.10} & \underline{28.77}
& \underline{41.58} & \underline{63.67} & \underline{42.29} & \underline{43.27} \\

\rowcolor{graybg} 100\%
& \textbf{60.24} & \textbf{35.13} & \textbf{68.81} & \textbf{42.35}
& \textbf{36.87} & \textbf{34.05} & \textbf{26.80} & \textbf{29.53}
& \textbf{42.32} & \textbf{64.21} & \textbf{43.01}
& \textbf{43.94} \\
\bottomrule
\end{tabular}%
}
\vskip -0.1in
\end{table*}

\begin{table*}[t]
\centering
\setlength{\tabcolsep}{2pt}
\scriptsize
\caption{Effect of instruction-tuning data scale. We vary the proportion of \method{}-471K-Instruction used for instruction tuning while keeping the full 1.2M pretraining data fixed.}
\label{tab7}
\renewcommand{\arraystretch}{1.18}
\resizebox{0.95\textwidth}{!}{%
\begin{tabular}{lcccccccccccc}
\toprule
\multirow{2}{*}{\textbf{SFT Ratio}}
& \multicolumn{4}{c}{\textbf{General}}
& \multicolumn{4}{c}{\textbf{Reasoning}}
& \multicolumn{3}{c}{\textbf{Hallucination}}
& \multirow{2}{*}{\textbf{Avg. $\uparrow$}} \\
\cmidrule(lr){2-5} \cmidrule(lr){6-9} \cmidrule(lr){10-12}
& \texttt{MME} & \texttt{MMStar} & \texttt{SQA} & \texttt{RealWorldQA}
& \texttt{MMMU} & \texttt{MMMU-P} & \texttt{VisuLogic} & \texttt{LogicVista}
& \texttt{CRPE} & \texttt{POPE} & \texttt{HallBench} & \\
\midrule
30\%
& 50.96 & 30.80 & 60.73 & 35.28
& 25.84 & 22.73 & 18.40 & 20.16
& 32.45 & 55.88 & 34.12 & 35.21 \\

60\%
& \underline{56.42} & \underline{33.40} & \underline{65.92} & \underline{39.15}
& \underline{31.76} & \underline{29.21} & \underline{23.60} & \underline{25.89}
& \underline{38.07} & \underline{61.53} & \underline{39.48} & \underline{40.40} \\

\rowcolor{graybg} 100\%
& \textbf{60.24} & \textbf{35.13} & \textbf{68.81} & \textbf{42.35}
& \textbf{36.87} & \textbf{34.05} & \textbf{26.80} & \textbf{29.53}
& \textbf{42.32} & \textbf{64.21} & \textbf{43.01}
& \textbf{43.94} \\
\bottomrule
\end{tabular}%
}
\end{table*}

\section{More Experiments} \label{App:E}

\subsection{Setting}

\paragraph{Training Setting.}
We build \method{} on top of Llama-3-8B-Instruct. The visual input is connected to the language model through a two-layer MLP projector with GELU activation. Instead of using image patch features during training, \method{} takes the synthetic image representations generated by our text-only pipeline as pseudo-visual tokens, and maps them into the LLM feature space through the projector. Training is conducted in two stages. In the first stage, we perform modality-alignment pretraining on \method{}-1.2M for one epoch, where only the projector is optimized, and the LLM backbone is frozen. The learning rate is set to $5\times10^{-4}$. In the second stage, we initialize the projector from the first stage and perform full-parameter instruction tuning on \method{}-471K-Instruction for one epoch, with the learning rate reduced to $1\times10^{-5}$. All experiments are run on 8 NVIDIA A100 GPUs. The complete training process takes approximately 27 hours.

\paragraph{Evaluation Setting.}
We evaluate \method{} on three groups of multimodal benchmarks. The general perception group includes MME \cite{fu2025mme}, MMStar\cite{chen2024we}, ScienceQA-Image\cite{lu2022learn}, and RealWorldQA. The complex reasoning group includes MMMU\cite{yue2024mmmu}, MMMU-Pro\cite{yue2025mmmu}, VisuLogic\cite{xu2025visulogic}, and LogicVista\cite{xiao2024logicvista}. The hallucination evaluation group includes CRPE\cite{wang2024all}, POPE\cite{li2023evaluating}, and HallusionBench\cite{guan2024hallusionbench}. For consistency across benchmarks, we report accuracy as the main evaluation metric.

\subsection{Scale Analysis}

\paragraph{Effect of Pretraining Scale.} Table~\ref{tab6} shows that Unicorn benefits consistently from larger text-only pretraining data. The average score increases from $40.24$ at 25\% data to $42.33$ at 50\%, $43.27$ at 75\%, and $43.94$ at 100\%. The gains exhibit diminishing returns, with larger improvement from 25\% to 50\% and smaller improvement from 75\% to 100\%, which is consistent with standard scaling behavior in pretraining. Importantly, the improvement appears across all benchmark groups, including general understanding, reasoning, and hallucination evaluation. This suggests that representation-level visual supervision is scalable: once text is made visually specific and transferred into the visual-compatible space, increasing text-only synthetic data continues to improve VLM performance.

\paragraph{Effect of SFT Scale.} Table~\ref{tab7} shows the effect of the instruction-tuning scale. Increasing the SFT ratio from 30\% to 60\% improves the average score from 35.21 to 40.40, and using the full SFT set further improves it to 43.94. Compared with the pretraining scale, the SFT scale has a larger impact on instruction-following and reasoning benchmarks, suggesting that sufficient instruction data is critical for adapting synthetic visual representations to downstream VQA-style interactions.


\section{Synthetic Data Examples}
\label{app:data_examples}

This section provides representative examples from the two synthetic datasets constructed by Unicorn. We show three examples from Unicorn-1.2M for pretraining and three examples from Unicorn-471K-Instruction for instruction tuning. Each pretraining sample consists of a dense visual caption and a captioning-style conversation, while each instruction-tuning sample contains multiple question-answer turns derived from the corresponding dense caption.

\subsection{Unicorn-1.2M}

\begin{tcolorbox}[promptbox,title=Pretraining Example 1]
\small
\textbf{ID:} 910010

\vspace{2pt}
\textbf{Caption.}
The person is standing next to a structure that is primarily composed of a metal frame, which is sleek and robust, with clean, straight lines and a polished finish that reflects ambient light. The frame supports an inset beveled glass top, which is rectangular in shape and has a slightly angled edge that creates a subtle, refined border. The glass is transparent and free of noticeable imperfections, allowing a clear view through its surface. The combination of the metal frame and beveled glass top gives the structure a modern and elegant appearance.

\vspace{4pt}
\textbf{Conversation.}

\vspace{2pt}
\textbf{Human:}
Provide a detailed and rich narrative explaining each component of the image. \texttt{<image>}

\vspace{2pt}
\textbf{GPT:}
The person is standing next to a structure that is primarily composed of a metal frame, which is sleek and robust, with clean, straight lines and a polished finish that reflects ambient light. The frame supports an inset beveled glass top, which is rectangular in shape and has a slightly angled edge that creates a subtle, refined border. The glass is transparent and free of noticeable imperfections, allowing a clear view through its surface. The combination of the metal frame and beveled glass top gives the structure a modern and elegant appearance.
\end{tcolorbox}

\begin{tcolorbox}[promptbox,title=Pretraining Example 2]
\small
\textbf{ID:} 910011

\vspace{2pt}
\textbf{Caption.}
A mature gum tree, approximately one meter in diameter at the base and with a smooth, grayish-bark, has fallen across a residential property. The tree's trunk and a significant portion of its branches have come to rest on both a person and the roof of a house. The person is partially obscured by the tree, with their lower body visible beneath the trunk, while the roof shows signs of damage from the impact, with shingles dislodged and a section of the structure visibly deformed. The surrounding area is cluttered with broken branches and leaves, and the ground is marked by the uprooted tree's exposed roots.

\vspace{4pt}
\textbf{Conversation.}

\vspace{2pt}
\textbf{Human:}
Provide a comprehensive and detailed description of the image, emphasizing narrative and visuals. \texttt{<image>}

\vspace{2pt}
\textbf{GPT:}
A mature gum tree, approximately one meter in diameter at the base and with a smooth, grayish bark, has fallen across a residential property. The tree's trunk and a significant portion of its branches have come to rest on both a person and the roof of a house. The person is partially obscured by the tree, with their lower body visible beneath the trunk, while the roof shows signs of damage from the impact, with shingles dislodged and a section of the structure visibly deformed. The surrounding area is cluttered with broken branches and leaves, and the ground is marked by the uprooted tree's exposed roots.
\end{tcolorbox}

\begin{tcolorbox}[promptbox,title=Pretraining Example 3]
\small
\textbf{ID:} 910012

\vspace{2pt}
\textbf{Caption.}
A blue and silver train, consisting of six carriages, is stationary on the platform at the railway station. The train's exterior is sleek, with a series of large, rectangular windows spaced evenly along the length of each carriage. The doors are aligned with the platform's edge, and a digital display above the doors shows the destination and next stop. The platform is well-lit by overhead lights, and yellow safety lines are painted on the ground to guide passengers. A few passengers are visible, standing near the train, some consulting their phones or waiting with luggage. The station's architecture includes a modern, glass canopy that extends over the platform, providing shelter from the elements.

\vspace{4pt}
\textbf{Conversation.}

\vspace{2pt}
\textbf{Human:}
\texttt{<image>} Offer a complete visual description of the photo with an emphasis on details.

\vspace{2pt}
\textbf{GPT:}
A blue and silver train, consisting of six carriages, is stationary on the platform at the railway station. The train's exterior is sleek, with a series of large, rectangular windows spaced evenly along the length of each carriage. The doors are aligned with the platform's edge, and a digital display above the doors shows the destination and the next stop. The platform is well-lit by overhead lights, and yellow safety lines are painted on the ground to guide passengers. A few passengers are visible, standing near the train, some consulting their phones or waiting with luggage. The station's architecture includes a modern, glass canopy that extends over the platform, providing shelter from the elements.
\end{tcolorbox}

\subsection{Unicorn-471M-Instruction}

\begin{tcolorbox}[promptbox,title=Instruction-Tuning Example 1: Multiple Choice]
\small
\textbf{ID:} 337637

\vspace{2pt}
\textbf{Caption.}
The beekeeper, dressed in a full protective suit with a white, mesh-covered helmet, reaches into the wooden hive box to extract a rectangular frame. The frame, approximately 18 inches long and 9 inches wide, is made of light-colored wood and is filled with hexagonal honeycomb cells. Each cell is meticulously constructed and some are filled with honey, while others contain developing bee larvae. The surface of the frame is covered with a dense layer of bees, their bodies varying in color from golden yellow to dark brown, with distinct black stripes. The bees are moving steadily, some clustering around the cells while others are engaged in various tasks. The beekeeper's gloved hands are carefully handling the frame to avoid disturbing the bees excessively. The hive box, constructed of weathered, light brown wood, is set on a wooden stand to elevate it off the ground. The surrounding area is a grassy field with scattered wildflowers, providing a natural habitat for the bees.

\vspace{4pt}
\textbf{Conversation.}

\vspace{2pt}
\textbf{Human:}
What is the beekeeper wearing to protect himself?

A: A light jacket and jeans

B: A full protective suit with a white, mesh-covered helmet

C: A raincoat and boots

D: A beekeeper's hat and gloves

\texttt{<image>}

\textbf{GPT:} B

\vspace{2pt}
\textbf{Human:}
What is the beekeeper doing with the frame?

A: Painting it

B: Extracting it from the hive box

C: Cleaning it

D: Installing it in a new hive

\textbf{GPT:} B

\vspace{2pt}
\textbf{Human:}
What is the approximate size of the frame the beekeeper is handling?

A: 12 inches long and 6 inches wide

B: 18 inches long and 9 inches wide

C: 24 inches long and 12 inches wide

D: 15 inches long and 7 inches wide

\textbf{GPT:} B

\vspace{2pt}
\textbf{Human:}
What are the hexagonal cells in the frame used for?

A: Storing pollen and nectar

B: Housing developing bee larvae and storing honey

C: Providing structural support

D: Protecting the bees from predators

\textbf{GPT:} B

\vspace{2pt}
\textbf{Human:}
What is the color of the wood used to construct the hive box?

A: Dark brown

B: Black

C: Weathered, light brown

D: White

\textbf{GPT:} C
\end{tcolorbox}

\begin{tcolorbox}[promptbox,title=Instruction-Tuning Example 2: Visual Question Answering]
\small
\textbf{ID:} 567214

\vspace{2pt}
\textbf{Caption.}
The serial number label on the back of a rectangular electronic device is prominently displayed. The device itself has a flat, matte surface with a uniform dark gray color. Its rectangular shape measures approximately 10 inches in length and 6 inches in width. The back panel is smooth and unadorned except for the label, which is affixed in the center. The label is a standard white sticker with black text, measuring about 2 inches by 1 inch. The text on the label is clearly printed and includes a series of alphanumeric characters, specifically ``SN1234567890ABCD.'' The label also features a small barcode below the serial number, which spans about 1.5 inches in length. The edges of the device are slightly rounded, and there are no visible screws or other fasteners on the back panel.

\vspace{4pt}
\textbf{Conversation.}

\vspace{2pt}
\textbf{Human:}
What is prominently displayed on the back of the device? \texttt{<image>}

\vspace{2pt}
\textbf{GPT:}
The serial number label is prominently displayed on the back of the device.

\vspace{2pt}
\textbf{Human:}
Why is the label placed in the center of the back panel?

\vspace{2pt}
\textbf{GPT:}
The label is placed in the center of the back panel to ensure it is easily visible and accessible for identification and reference.

\vspace{2pt}
\textbf{Human:}
How does the device's surface feel to the touch?

\vspace{2pt}
\textbf{GPT:}
The device's surface feels flat and matte, with a uniform dark gray color.

\vspace{2pt}
\textbf{Human:}
Does the back panel have any visible screws or fasteners?

\vspace{2pt}
\textbf{GPT:}
No, the back panel is smooth and unadorned except for the label, and there are no visible screws or other fasteners.

\vspace{2pt}
\textbf{Human:}
Is the serial number label a standard white sticker?

\vspace{2pt}
\textbf{GPT:}
Yes, the serial number label is a standard white sticker with black text.

\vspace{2pt}
\textbf{Human:}
Where is the barcode located on the label?

\vspace{2pt}
\textbf{GPT:}
The barcode is located below the serial number on the label.

\vspace{2pt}
\textbf{Human:}
Who might use the serial number and barcode on the device?

\vspace{2pt}
\textbf{GPT:}
The serial number and barcode might be used by manufacturers, service centers, and customers for tracking, warranty claims, and product identification.

\vspace{2pt}
\textbf{Human:}
Which dimensions does the label have?

\vspace{2pt}
\textbf{GPT:}
The label measures about 2 inches by 1 inch.
\end{tcolorbox}

\begin{tcolorbox}[promptbox,title=Instruction-Tuning Example 3: Complex Reasoning]
\small
\textbf{ID:} 407133

\vspace{2pt}
\textbf{Caption.}
A boy, approximately 8 years old, is running in shallow water at a beach. He is wearing blue and green swim trunks with a pattern of small waves. His skin is lightly tanned, and his hair, which is light brown, is damp and slightly tousled from the water. His arms are swinging at his sides in a natural running motion, and his legs are spread apart, with one foot just breaking the surface of the water and the other submerged, kicking up a small splash. The water around him is clear and calm, with a light ripple from his movement. The beach sand is visible beneath the water, and the horizon shows a clear blue sky with a few fluffy white clouds.

\vspace{4pt}
\textbf{Conversation.}

\vspace{2pt}
\textbf{Human:}
Based on the image caption, what can you infer about the weather conditions at the beach? \texttt{<image>}

\vspace{2pt}
\textbf{GPT:}
The weather conditions at the beach can be inferred to be sunny and pleasant. The caption mentions a clear blue sky with a few fluffy white clouds, which indicates a sunny day with minimal cloud cover. This type of weather is typically associated with warm and comfortable temperatures, ideal for beach activities like swimming and running in the water.

\vspace{2pt}
\textbf{Human:}
What evidence in the caption suggests that the boy has been in the water for some time?

\vspace{2pt}
\textbf{GPT:}
The caption provides several pieces of evidence suggesting that the boy has been in the water for some time. His hair is described as damp and slightly tousled from the water, indicating that he has been in the water recently and possibly for a while. Additionally, his skin is lightly tanned, which could suggest he has been exposed to the sun for a period, likely while engaging in water activities.

\vspace{2pt}
\textbf{Human:}
If the boy is running in shallow water, what can you deduce about the depth of the water and the safety of the beach?

\vspace{2pt}
\textbf{GPT:}
The caption states that the boy is running in shallow water, which suggests that the water is not deep enough to pose a significant risk of drowning. The fact that he can run and his legs are spread apart, with one foot breaking the surface and the other submerged, indicates that the water is likely only a few inches to a foot deep. This shallow depth, combined with the clear and calm water and visible sand beneath, suggests that the beach is relatively safe for children to play in.
\end{tcolorbox}


\section{Qualitative Analysis}
\label{sec:qualitative}

\begin{figure*}[h!]
  \centering
  \includegraphics[width=0.9\textwidth]{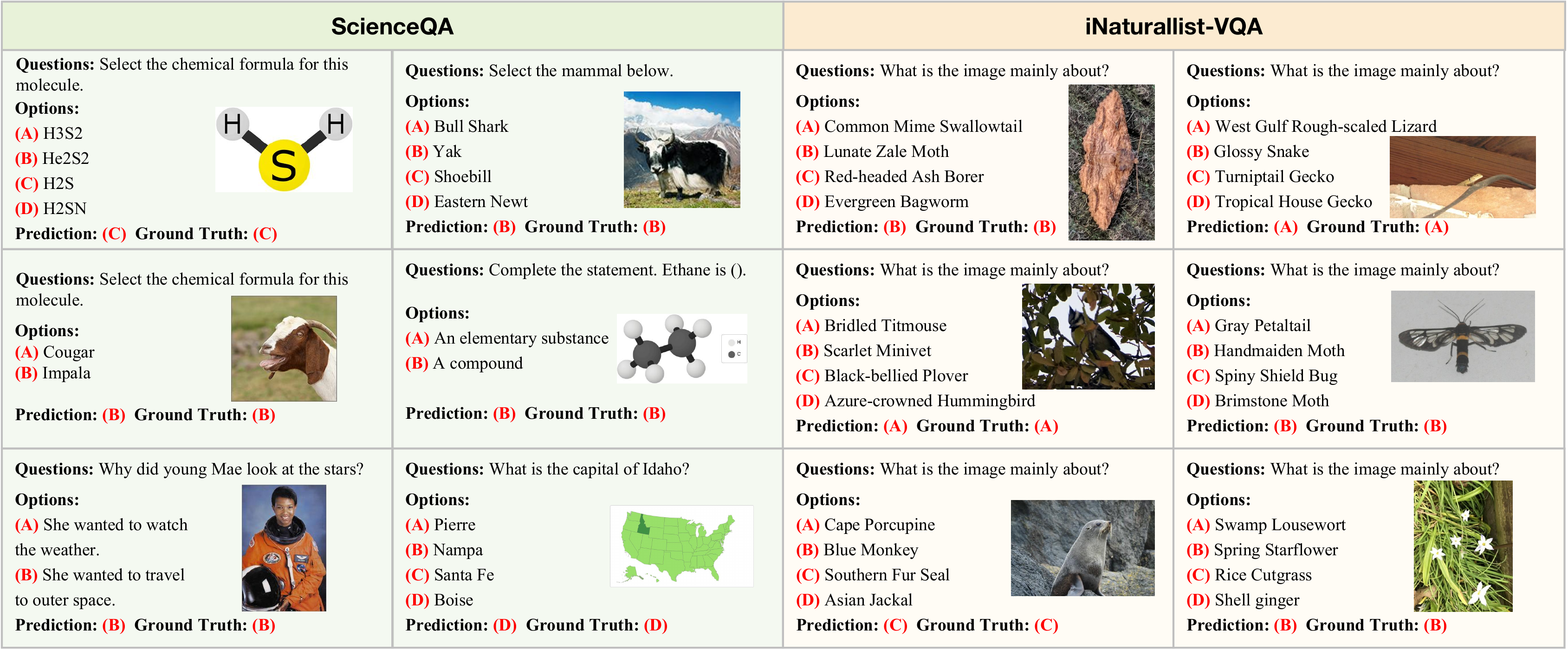}
  \caption{Examples of Unicorn-8B’s correct responses on the ScienceQA and iNaturallist-VQA.}
  \label{fig:qualitative}
\end{figure*}

Fig.~\ref{fig:qualitative} presents qualitative examples of \method{} on two types of visual question-answering tasks. The left part shows examples from ScienceQA, while the right part shows examples from our fine-grained iNaturalist-VQA test set. 


\paragraph{ScienceQA.} The ScienceQA examples cover different forms of visual understanding, including chemical structure recognition, object/category identification, commonsense reasoning, and map-based knowledge reasoning. For instance, \method{} correctly identifies the molecular formula from a visual chemical structure, selects the mammal from candidate animals, infers why a child looks at the stars based on the astronaut context, and answers a geography question using the map cue. These cases suggest that the model is not merely relying on language priors, but can integrate visual evidence with textual options to produce the correct answer.

\paragraph{iNaturalist-VQA.} To further evaluate fine-grained visual recognition, we construct iNaturalist-VQA, a fine-grained animal and plant VQA test set. Compared with general VQA examples, this setting is more challenging because the candidate categories often belong to visually similar biological groups, requiring the model to capture subtle visual cues such as body shape, texture, color patterns, and local structures. As shown in Fig.~\ref{fig:qualitative}, \method{} correctly distinguishes categories such as moths, lizards, birds, marine mammals, and plants. These results indicate that the synthetic supervision produced by Unicorn preserves useful fine-grained visual semantics and can support recognition beyond coarse object categories.


\end{document}